\documentclass[11pt,a4paper]{article}

\usepackage{times,latexsym}
\usepackage{url}
\usepackage[T1]{fontenc}
\usepackage{array,ragged2e}
\usepackage{setspace}

\newcolumntype{P}[1]{>{\RaggedRight\arraybackslash}p{#1}}
\usepackage{caption}
\DeclareCaptionFont{myfont}{\fontsize{10pt}{10pt}\selectfont}

\usepackage[acceptedWithA]{tacl2021v1}
\usepackage{comment}
\usepackage{xspace,mfirstuc,tabulary}

\newif\iftaclinstructions
\taclinstructionsfalse % AUTHORS: do NOT set this to true
\iftaclinstructions
\renewcommand{\confidential}{}
\renewcommand{\anonsubtext}{(No author info supplied here, for consistency with
TACL-submission anonymization requirements)}
\newcommand{\instr}
\fi

\iftaclpubformat % this "if" is set by the choice of options

\else

\fi

\title{Modelling Analogies and Analogical Reasoning: Connecting Cognitive Science Theory and NLP Research.}

\author{Molly R. Petersen\textsuperscript{1,2} \and Claire E. Stevenson\textsuperscript{3} \and Lonneke van der Plas\textsuperscript{1,4} \\
  \textsuperscript{1} Computation, Cognition \& Language Group, Idiap Research Institute, Martigny, Switzerland \\
  \textsuperscript{2} NLP Lab, EPFL,  Lausanne, Switzerland \\
    \textsuperscript{3} Psychological Methods, University of Amsterdam, the Netherlands\\
 \textsuperscript{4} Faculty of Communication, Culture and Society and Faculty of Informatics\\ Università della Svizzera italiana, Lugano, Switzerland\\
  \texttt{molly.petersen@epfl.ch}, \texttt{c.e.stevenson@uva.nl}, \texttt{lonneke.vanderplas@usi.ch}}

\date{}

\begin{document}

\maketitle
\begin{abstract}
Analogical reasoning is an essential aspect of human cognition. In this paper, we summarize key theory about the processes underlying analogical reasoning from the cognitive science literature and relate it to current research in natural language processing. While these processes can be easily linked to concepts in NLP, they are generally not viewed through a cognitive lens. Furthermore, we show how these notions are relevant for several major challenges in NLP research, not directly related to analogy solving.  This may guide researchers to better optimize relational understanding in text, as opposed to relying heavily on entity-level similarity.

\end{abstract}

\section{Introduction}

Relational reasoning- and by extension analogy and analogical reasoning- has always held a prominent place in human psychology. Some have argued that analogy is central to the human cognitive experience \cite{hofstadter2001epilogue}. The fact that analogies were used to demonstrate the emerging capabilities of the representations produced by Word2Vec \cite{mikolov2013efficientestimationwordrepresentations, mikolov2013distributedrepresentationswordsphrases} when it was first introduced, is perhaps evidence that underscores the value we place on analogy and relational reasoning to understanding our world and the knowledge contained in it.

In cognitive science, perhaps the most well-known theory of analogical reasoning is Structural Mapping Theory (SMT) introduced by \citet{gentner1983structure}. Here, Gentner differentiates types of similarity between two systems- a source concept and target concept (which can be thought of as a domain and codomain), for example between the source concept \emph{solar system} and target concept \emph{atom}. Similarity can be measured along two axes: \emph{attribute similarity} between entities (ex: comparing the size difference between a planet and an electron), and \emph{relational similarity}, the extent relations between entities or predicates within each system are similar across systems (ex: a planet orbits the sun in our solar system, and an electron orbits a nucleus in an atom). \emph{Literal similarity} holds when both attribute and relational similarity between two systems is high, \emph{mere appearance matches} when attribute similarity is high while relational similarity is low, and \emph{analogy}, when attribute similarity is low but relational similarity is high. She also introduces the \emph{systematicity principle}, where preference is given to preserving higher-order relations between two systems.

 Because of the extent of attribute and relational similarity between two systems is a continuum, the extent to which a comparison between any two systems fits within these categories is also a continuum, and depends on the context of the comparison taking place \cite{gentner1987mechanisms}. An electron and planet may both share the attribute of roundness, however they are dissimilar in many other attribute dimensions that neither mere appearance or literal similarity can accurately describe. Where they are similar is through their relations to other entities in their respective systems.

Regardless of whether one thinks that analogy is merely a common, useful tool for human cognition, or that it is at the core of it, the role it plays in cognition is profound and therefore the goal of making models smarter and more human-like should incorporate this type of reasoning.

In this paper, we cover key theory from the cognitive science literature regarding analogical reasoning, and review recent research regarding analogy solving in natural language processing (NLP). The goal of this paper is not to provide a detailed compilation of all analogical reasoning research in both the fields of cognitive science and computer science, but instead create an introduction for NLP researchers. We focus specifically on the processes of analogical reasoning, each of which have a large body of research, as well as their own unresolved questions. We chose to focus on this aspect of analogical reasoning since most of these processes are more or less already defined concepts in NLP. However, in practice they are not often viewed through a cognitive lens, but purely by their definitions in the computer science field. We also limit our discussion to computational methods of NLP using neural word embeddings, as reviews of other methods have already been published elsewhere \cite{holyoak1989computational,gentner2011computational,Mitchell_2021}. Since our focus is on NLP, we focus only on text-based analogies.

 Lastly we argue that analogical reasoning is highly relevant to NLP at large, and can potentially be a way in which common limitations with current NLP models can be addressed. We do not provide specific answers to these problems, nor do we think analogical reasoning is the ultimate answer to addressing these problems. We instead wish to create a starting point for the NLP research community to engage in research that is theoretically motivated by the cognitive science literature.
 
 Our paper is organized as follows- \textbf{Section \ref{cogsci}:}, we summarize theory from cognitive science regarding the analogical reasoning processes in humans, \textbf{Section \ref{level}:} we discuss different types of analogies as well as their specific considerations, \textbf{Section \ref{nlp}:} we summarize recent work in analogy solving and more broadly analogical reasoning in NLP, connecting them to the analogical reasoning processes discussed in Section 2, and finally, \textbf{Section \ref{uses}:} we discuss domain-specific applications for analogies, as well as re-frame common NLP tasks and challenges through an analogical lens.

\section{Analogical Reasoning in Cognitive Science} \label{cogsci}

In the cognitive science literature, analogical reasoning is split into different processes. These are 1) {\fontsize{11pt}{11pt}\selectfont\textsc{retrieval}}, 2) {\fontsize{11pt}{11pt}\selectfont\textsc{mapping}}, 3) {\fontsize{11pt}{11pt}\selectfont\textsc{representation}}, 4) {\fontsize{11pt}{11pt}\selectfont\textsc{abstraction}}, and 5) {\fontsize{11pt}{11pt}\selectfont\textsc{encoding}}. The literature will sometimes vary on which processes are included \cite{gentner1987mechanisms,holyoak1989analogical,gentner1993roles,thagard1990analog,gentner2002learning,kokinov2003computational,gentner2011computational,gentner2012humanbehavior}, however {\fontsize{11pt}{11pt}\selectfont\textsc{retrieval}} and {\fontsize{11pt}{11pt}\selectfont\textsc{mapping}} are always discussed. {\fontsize{11pt}{11pt}\selectfont\textsc{Mapping}} specifically is considered the central process of analogy making \cite{gentner2011computational}. 

Additionally, there is often a distinction made between near and far analogies- near analogies being when both systems of an analogy come from a similar domain (and therefore the elements inside it are likely semantically similar), and far analogy being where they come from two disparate domains \cite{barnett2002and}. It has been demonstrated in humans that solving near analogies is generally easier than far analogies \cite{bunge2005analogical, jones2022differential}. However, far analogies have been demonstrated to better encourage novel problem solving and promote relational reasoning \cite{cagan2011effective,vendetti2014far,chan2015impact, walker2018achieving}.

In this section, we will briefly discuss all five processes, and use the example of \emph{planet:sun::electron:nucleus}, where the single \emph{:} stands for \emph{orbits}, to more concretely explain the processes. It's important to note that these processes are not necessarily independent of each other or performed step-by-step. They can be done simultaneously and are interdependent \cite{holyoak1987surface}, therefore the order we present them in does not hold any special significance other than the ease of explanation. The processes and examples are summarized in Table \ref{tab:table1}.

\begin{table*}[t!]

\begin{center}

\caption{Summarization of the processes of analogical reasoning using the example \emph{planet:sun::electron:nucleus} and the relation \emph{orbits}.} 
\label{tab:table1}

\begin{tabular}{P{.20\textwidth}P{.74\textwidth}}
 \hline

 \multicolumn{1}{c}{\textbf{Process}} & \multicolumn{1}{c}{\textbf{Example}}\\
\hline\hline

 {\fontsize{11pt}{11pt}\selectfont\textsc{Retrieval}} & When given the pair \emph{planet:sun}, collecting potential matches from memory that may contain the \emph{revolve around} relation, such as \emph{electron:nucleus}, \emph{satellite:Earth}, and \emph{the World:you} \\ 
  \hline
 {\fontsize{11pt}{11pt}\selectfont\textsc{Mapping}} & Identifying that \emph{planet} plays the same role as \emph{electron} in the solar system and an atom, respectively\\ 
  \hline

{\fontsize{11pt}{11pt}\selectfont\textsc{Rerepresentation}} & Altering the representation of each system to improve the analogy. One does not need to represent each planet in the solar system or each electron in an atom individually, these entities can be grouped \\
 \hline
 {\fontsize{11pt}{11pt}\selectfont\textsc{Abstraction}} & Representing the solar system as a \emph{central force system}, where the particular entities in the general concept of a central force system are not specified \\
\hline

 {\fontsize{11pt}{11pt}\selectfont\textsc{Encoding}} & How an instance is represented in memory. Understanding the solar system as a \emph{central force system} vs. a \emph{gravitationally-bound system} \\ 

 \hline
\end{tabular}
\end{center}
\end{table*}

\subsection{Retrieval}\label{retrieve}
In an ideal scenario, one may be given a relevant example and told to map it to a pre-specified target domain, or pre-defined list of options to select an analogy from, therefore eliminating the need for {\fontsize{11pt}{11pt}\selectfont\textsc{retrieval}} to begin with. More typically, when presented with a scenario, in order to form a useful analogy one must retrieve an instance from memory that is an appropriate analogue \cite{holyoak1987surface}. In humans, this memory generally contains past experiences (episodic memory) and gained knowledge (semantic memory). For models, this would be the training data the model has seen, and occasionally also a knowledge base the model can query.

For an explicit example, when presented with the terms \emph{planet:sun}, a cognitive system must explore their knowledge base to retrieve either \emph{electron:nucleus} or another relevant pair. Notably, there are other possible pairs that could be retrieved which will vary on their validity and utility for any given use case. In theory other potential candidates could be \emph{satellite:earth} or \emph{the World:you}. Whether potential matches are appropriate for the specific use case is handled during evaluation.

{\fontsize{11pt}{11pt}\selectfont\textsc{Retrieval}} poses a large challenge within the analogical reasoning process. Not only does the pool of candidates for selection include all knowledge or experience present in the given cognitive system, but there is also the issue of how to retrieve a relevant instance when potential relevant instances may have little or no semantic or surface-level similarity to the target domain.  It has been shown in human studies that surface similarity effects people's abilities to retrieve relevant analogs despite people's preference for relational similarity in analogies \cite{gentner2003learning,gentner1993roles,holyoak1987surface,minervino2024surface}. This has been attributed to how information is processed in both the {\fontsize{11pt}{11pt}\selectfont\textsc{abstraction}} and {\fontsize{11pt}{11pt}\selectfont\textsc{encoding}} steps, which we describe later \cite{gentner2003learning}.

\subsection{Mapping}\label{map}

{\fontsize{11pt}{11pt}\selectfont\textsc{mapping}} is considered the crux of analogical reasoning- without {\fontsize{11pt}{11pt}\selectfont\textsc{mapping}} between two systems there is no analogy. In our solar system example, this would be {\fontsize{11pt}{11pt}\selectfont\textsc{mapping}} the entity \emph{planet} to \emph{electron} and \emph{sun} to \emph{nucleus} by noticing that they play the same roles in relation with \emph{orbits}. One can use this initial mapping to perform additional mappings between the two systems, such as connecting the role of \emph{gravity} to that of \emph{electromagnetism}. As stated previously, it is difficult to discuss
analogy without mentioning Gentner’s SMT \cite{gentner1983structure} outlined in the introduction, however many people have expanded on or proposed alternatives to her original theory.

\citet{holyoak1989analogical} introduced three {\fontsize{11pt}{11pt}\selectfont\textsc{mapping}} constraints gathered from common themes that appear in the analogy literature. The authors specify that these are not requirements, but instead act as "pressures" to be optimized during the analogical reasoning process. The first constraint is \emph{isomorphism}, or the idea that mappings should be one-to-one and complete (bijective). This forms a structural constraint between two systems. Ideally matches are one-to-one and complete, but this is only a constraint and not an absolute criterion. So, not all components must be matched, and one-to-many and many-to-one matches are also possible. This can partially be addressed in the {\fontsize{11pt}{11pt}\selectfont\textsc{rerepresentation}} process covered in Section \ref{rerep}. The second constraint is that of \emph{semantic similarity}, where the similarity in meaning between entities in the two systems should be considered to some extent. 

Lastly, they argue that purpose should guide the {\fontsize{11pt}{11pt}\selectfont\textsc{mapping}} process with the constraint of \textit{pragmatic centrality}. Analogical reasoning and {\fontsize{11pt}{11pt}\selectfont\textsc{mapping}} always take place in the context of an overarching goal or purpose, and all steps of analogical reasoning, {\fontsize{11pt}{11pt}\selectfont\textsc{mapping}} included, will depend on the intended purpose and must take that into account. In theory, a large number of components in two systems can be mapped, or there could be different combinations of mappings between two systems, however, not all are universally relevant to all use cases.

\subsection{Rerepresentation}\label{rerep}
{\fontsize{11pt}{11pt}\selectfont\textsc{rerepresentation}} (not to be confused with \emph{representation} as typically used in NLP, which is probably closest to the {\fontsize{11pt}{11pt}\selectfont\textsc{encoding}} process described in section \ref{abtract}) is the process of altering a representation of either or both the source or target domain to improve a match between the two systems \cite{gentner2011computational}. Knowledge regarding a particular system can change over time due to experience or education, requiring updates to how it is represented in memory \cite{yan2013theory}. 

\citet{yan2013theory} defined several methods to assist {\fontsize{11pt}{11pt}\selectfont\textsc{rerepresentation}}: \emph{truth-preserving transformation} to the structure (e.g. a planet being smaller than the sun can also be expressed as the sun being larger than a planet); \emph{decomposition} of relations which still preserves the relevant features of the original relations (e.g. electrons do not have elliptical orbits like planets, but this feature may not be relevant to the analogy); \emph{entity collecting}, grouping entities when multiple entities play equivalent roles (our solar system has eight planets, which can be grouped when comparing to electrons); and \emph{entity splitting}, which helps one-to-one {\fontsize{11pt}{11pt}\selectfont\textsc{mapping}} when a particular entity plays multiple roles by splitting these objects into individual parts.

\subsection{Abstraction and Encoding}\label{abtract}
In the original SMT paper, {\fontsize{11pt}{11pt}\selectfont\textsc{abstraction}} was defined as a type of similarity separate from analogy, however in more recent papers {\fontsize{11pt}{11pt}\selectfont\textsc{abstraction}} has been considered part of the process of analogical reasoning \cite{gentner1983structure,gentner2017analogy,gentner2011computational}. \citet{gentner2017analogy} defines {\fontsize{11pt}{11pt}\selectfont\textsc{abstraction}} as "decreasing the specificity (and thereby increasing the scope) of a concept" (pg. 673). In analogy, the focus is generally on abstractions of relations, creating rules or frameworks of a general concept from more specific examples. Given that surface similarity generally corresponds to specific details between two systems, {\fontsize{11pt}{11pt}\selectfont\textsc{abstraction}} can then help generalize and decrease the influence of irrelevant and distracting details \cite{gick1980analogical}.

{\fontsize{11pt}{11pt}\selectfont\textsc{encoding}} is how a particular structure is represented in memory. This can be a single specific example or the schema resulting from {\fontsize{11pt}{11pt}\selectfont\textsc{abstraction}} \cite{mandler1993analogical}. How an abstraction or instance is encoded has been suggested as the key to effective {\fontsize{11pt}{11pt}\selectfont\textsc{retrieval}} from memory as well as analogical transfer \cite{gick1983schema,loewenstein2010one, mandler1993analogical}.

For example, both a solar system and an atom can be abstracted as a central force system \cite{gentner1983structure}. In each case, the object of central force differs, as does the orbiting object and the force creating the orbit itself. The specifics of the objects that play each role has been abstracted away, they are slots to be filled with specific entities. How this is encoded in memory will have an effect on how it can be retrieved and compared later- if the solar system was encoded as a gravitationally-bound system in memory, it may be harder to retrieve an atom as an appropriate analogy \cite{gick1983schema}. Additionally, it can be encoded in memory as both, however which encoding is associated and retrieved can vary given the specific context \cite{gentner2004analogical}, and subsequently effects analogical transfer.

One potential challenge with {\fontsize{11pt}{11pt}\selectfont\textsc{encoding}} and {\fontsize{11pt}{11pt}\selectfont\textsc{abstraction}} is word choice and how that affects a cognitive system's perception of the information that is contained in text, or more generally, how information is presented to optimize extraction of relevant details \cite{yan2013theory,ramscar2003semantic,gentner2003learning,gick1980analogical}. Additionally, there is the limitation of how much knowledge a cognitive system contains in order to formulate a representation that includes relevant details that can be mapped \cite{yan2013theory,hummel1997distributed}. These challenges can be addressed in the {\fontsize{11pt}{11pt}\selectfont\textsc{rerepresentation}} stage \cite{forbus1998analogy,gentner2003learning}.

\section{Problem Types}\label{level}

We will now describe our classification of analogy types and discuss their specific considerations. There has been other work defining a taxonomy of different analogy types \cite{thilinilevels, wijesiriwardene2023analogical, nagarajah2022understandingnarrativesdimensionsanalogy}. For example, \citet{thilinilevels} presents a taxonomy of analogies based on the depth of knowledge and information a system needs. For the purpose of this paper, we define analogies as belonging to one of three types: symbolic, entity-level, or contextualized. This categorization is not meant to compete with or replace other taxonomies, but instead to define and discuss three broad categories based on data format, which determine the approaches NLP researchers may choose to handle them. 

\subsection{Symbolic Analogies}

First, we discuss analogies that, while made up of text, generally contain no semantic information. They are strings of letters and symbols that contain patterns which are solvable by humans \cite{hofstadter1995copycat}. For example, given \emph{abc $\rightarrow$ abd}, finding the transformation for \emph{lmn}. Additionally, this type of analogy could also include {\fontsize{11pt}{11pt}\selectfont\textsc{mapping}} words in natural language to non-semantic strings of characters \cite{musker2024semanticstructuremappingllmhuman} as well as digit matrices (for example, a 3 x 3 matrix with a blank at the position at [3,3] which needs to be filled by the solver)  \cite{webb2023emergentanalogicalreasoninglarge}.

One challenging aspect of this type of analogy is the potential to apply a literal, but valid, rule to solve the analogy. For example, for \emph{abc $\rightarrow$ abd}, the conceptual rule is to change the last letter in the sequence to the next letter in the alphabet, resulting in \emph{lmo}. However, a potential response could be \emph{lmd}, where the assumed rule would be to replace the last letter with \emph{d}, regardless of the original letter \cite{lewis2024counterfactual}.

Research on symbolic analogies in NLP is generally less prevalent than the other categories of analogy, perhaps because as mentioned before they often do not involve semantic information, and by extension the ability to solve these sorts of analogies are probably not indicative of any sort of natural language understanding. They do, however, function as a tool to measure reasoning and pattern recognition.

\subsection{Entity-Level Analogies}

Also called proportional analogies, these are analogies that are expressed in the format \emph{a:b::c:d}, verbally expressed as \emph{a} is to \emph{b} as \emph{c} is to \emph{d}. These can be extended to analogies that contain more than 4 entities (ex \emph{planet:sun:gravity::electron:nucleus:electromagn-etism}). However, they do not contain additional language that contextualizes the entities in a larger picture, or extra textual information that models would need to either consider or discard to identify the analogy. Solving these analogies involves either selecting the best option from a predefined list of potential answers, or generating an answer from existing knowledge.

\subsubsection{Morpho-syntactic Analogies}
 In NLP, the most well known English language benchmarks for analogy are probably the Google Analogy Test Set \cite{mikolov2013efficientestimationwordrepresentations} and the Bigger Analogy Test Set (BATS) \cite{gladkova-etal-2016-analogy}. These datasets are largely comprised of analogies where the "is to" relation is a morphological relation such as pluralizing a singular noun (ex: \emph{member:members::fact:facts}). Versions of these analogy datasets have been released for a variety of languages \cite{ulčar2020multilingualcultureindependentwordanalogy,KarpinskaLiEtAl_2018,krishnan-ragavan-2021-morphology,grave-etal-2018-learning}.

While morphological analogy benchmarks fit under the definition of analogy, they are limited by the fact they aren't very representative of human analogical reasoning \cite{ushio-etal-2021-bert, petersen-van-der-plas-2023-language}. They are trivially easy to solve- if given the pair \emph{try:trying} and then asked to complete the analogy with the made up verb \emph{"zoop"}, the fourth term is clearly \emph{"zooping"}.
 
 Here \emph{"zoop"} has no definition, nor is knowing the meaning of \emph{"zoop"} required to solve the analogy. That is not to say these analogies have no use, there are potential creative uses of being able to correctly conjugate novel words or morph already existing words into different parts of speech. But the relational reasoning that can be tested with these benchmarks is ultimately limited \cite{ushio-etal-2021-bert}.
 
\subsubsection{Semantic Analogies}

 Datasets in this category include the Scientific and Creative Analogy dataset (SCAN) \cite{czinczoll2022scientificcreativeanalogiespretrained}, analogies created to test human analogical reasoning (termed psychometric analogies by \citet{ushio-etal-2021-bert}) such as the SAT dataset \cite{turney2003learning} and the Knowledge-intensive Analogical Reasoning benchmark (E-KAR) \cite{Chen_2022}, as well as some relations present in the Google Analogy Testset and BATS (e.g. encyclopedic semantics). With these analogies, one must understand all terms present in the analogy, as well as how they relate to each other.

\subsubsection{Vocabulary Beyond the Layperson's Dictionary}

 While entity-level analogies may be thought of as containing words that could be found in a dictionary compiled for use by the everyday person, this concept can be extended and applied in more creative ways to include domain-specific jargon and concepts. For example \citet{yamagiwa2024predictingdruggenerelationsanalogy} used this level of analogy for drug-gene relations where the entities consisted of drugs and genes (ex: \emph{bosutinib:ABL1}). \citet{blair2021ai} used entities to represent U.S. tax codes and concepts, in which different tax codes form analogies by pertaining to similar rules, regulations, topics etc.

%\cite{yuan-etal-2024-analogykb}
\subsubsection{Entity-Level Analogies- Considerations}

As mentioned previously, solving these analogies involve either selection or generation, where generation includes the {\fontsize{11pt}{11pt}\selectfont\textsc{retrieval}} process while selection does not. If the goal is to generate an answer, arguably many words can be correct and vary in degree of correctness, creating a challenge for evaluation \cite{rogers-etal-2017-many}.

The lack of context poses a challenge with these analogies. Words can be polysemous, and the precise definition employed by an entity may not be obvious given the other entities. If there is no other entity in the concept, as will often be the case in analogies of the format \emph{a:b::c:?}, the meaning has to be inferred from the source concept. The effect of lack of context will depend on the part of speech, word frequency, and other qualities of a particular entity \cite{gentner1988verb, asmuth2017relational, fenk2010frequency}. These issues could potentially be addressed with the \emph{decomposition} method for {\fontsize{11pt}{11pt}\selectfont\textsc{rerepresentation}}.

It's known that analogies are permutable  (for example permuting \emph{a:b::c:d} to \emph{a:c::b:d}) \cite{marquer2022transferring, antic2022analogical}, however these permutations do not hold for every analogy. Given the analogy \emph{Lima:Peru::Cuzco:Peru}, where the relation is \emph{"city in"}, permuting the analogy to \emph{Lima:Cuzco::Peru:Peru}, clearly creates an invalid mapping, as \emph{Peru} = \emph{Peru}, while \emph{Lima} $\neq$ \emph{Cuzco}. Many may be non-injective functions, such as the example with Peruvian cities, for which these permutations can potentially create invalid mappings.  

\subsection{Contextual Analogies}

These are analogies that involve lengths of text ranging from phrases to passages. In line with the entity-level analogy between the solar system and an atom previously mentioned, comparing paragraphs describing these two systems, or even comparing whole Wikipedia pages between these two systems to find the analogical components, would fit into this category.

This category is obviously much more broad then entity-level analogies, and encompasses most use cases of analogies that you find "in the wild", including narratives \cite{gick1980analogical, sourati2024arnanalogicalreasoningnarratives,nagarajah2022understandingnarrativesdimensionsanalogy}, procedural texts \cite{sultan2023lifecircusclownsautomatically}, and legal texts \cite{santosh2024mindneighboursleveraginganalogous}.

\subsubsection{Contextual Analogies- Considerations}

While the lack of context may present challenges with entity-level analogies, the presence of context can also pose a challenge. The text may include information that is irrelevant to an analogy, or that distracts from the analogy with surface-level similarities or differences. It may also leave room to rely on spurious correlations, which may hinder {\fontsize{11pt}{11pt}\selectfont\textsc{abstraction}} and generalization \cite{gentner1991language, wang-culotta-2020-identifying}.
 
In line with the theory that analogy should take into account \emph{pragmatic centrality}, getting the model to consider the goal when making decisions could be challenging. When given text, it is not necessarily obvious what entities should be included in the {\fontsize{11pt}{11pt}\selectfont\textsc{mapping}} process, which may change based on the objective. This is opposed to entity-level analogies where maximizing the mappings between all provided entities will generally be the goal. Additionally, the objective will define what instance should be retrieved from memory.

\section{Current Research on Analogies in NLP}\label{nlp}

Given the role of analogies as benchmarks in NLP as well as their use as a measure of reasoning abilities in humans, the question naturally arises whether any aspect of analogical reasoning or types of analogies are considered "solved" in NLP. To some extent this is a difficult question to answer, as high performance on currently available benchmarks raises the question of whether or not these specific benchmarks have been seen in a model's pre-training data \cite{hodel2024responseemergentanalogicalreasoning, deng-etal-2024-investigating}.

Generally speaking, results from recent models suggest that analogies at any level are not solved, with the exception of morpho-syntactic analogies, where some models and experiments have demonstrated almost perfect accuracy \cite{chan:hal-03674913, yuan-etal-2024-analogykb}.

For example, after the introduction if GPT-3 \cite{brown2020language}, \citet{webb2023emergentanalogicalreasoninglarge} evaluated the model on a variety of analogy tasks (including tasks from all categories of analogy mentioned in the previous section), and found that GPT-3 performed the same or better than humans on a variety of analogy tasks making the claim that GPT-3 has the capacity to perform general reasoning tasks in the zero-shot setting. However, a response from \citet{hodel2024responseemergentanalogicalreasoning} criticized the ability of their experimental setup to evaluate general, zero-shot reasoning in GPT-3, mentioning the potential that the analogies used as tests where prevalent in the training data. The analogies they used have generally been around for quite a long time, such as Dunker's radiation problem \cite{duncker1945problem}, which was originally published over half a century ago and has been cited over 6000 times, and Hofstadter's copycat problems \cite{hofstadter1995copycat}, which were published in the 90's. \citet{hodel2024responseemergentanalogicalreasoning} found that models failed when the original tasks were modified, and that the model could accurately describe and provide an example to the copycat problem when prompted.

We will now cover specific approaches various researchers in NLP have taken with analogy tasks and language models (LMs). While computational modeling of analogies has been done for decades \cite{falkenhainer1989structure, holyoak1989analogical, thagard1990analog, hofstadter1995copycat, hummel1997distributed, doumas2008theory}, we focus on more recent advances of modeling analogies using neural word embeddings. We attempt to connect specific experiments to the processes outlined in  section \ref{cogsci}. In general, we find most research with entity-level analogies focuses on the {\fontsize{11pt}{11pt}\selectfont\textsc{mapping}} process. For contextual analogies, while motivating their work with theory from analogical {\fontsize{11pt}{11pt}\selectfont\textsc{mapping}}, notably, these works often do not explicitly perform any {\fontsize{11pt}{11pt}\selectfont\textsc{mapping}} between entities or relations, but instead perhaps more closely address the {\fontsize{11pt}{11pt}\selectfont\textsc{abstraction}} and {\fontsize{11pt}{11pt}\selectfont\textsc{encoding}} processes.

\subsection{Symbolic Analogies}

\citet{lewis2024counterfactual} tested several GPT models on symbolic analogies with the English alphabet, as well as with novel alphabets (permuted English alphabets and a symbolic alphabet), and digit matrices. They found that while humans could generally cope with alternative alphabets, GPT models struggled with them more compared to the standard alphabet. This suggests an issue with {\fontsize{11pt}{11pt}\selectfont\textsc{Retrieval, Abstraction}} and {\fontsize{11pt}{11pt}\selectfont\textsc{Rerepresentation}}, as they are not robust to reasoning over novel alphabets despite seeing alphabets and likely letter string analogies with the standard English alphabet in their training data. Additionally, on the digit matrices, human performance did not change much if the blank of the matrix was moved from the [3,3] position to an alternate position, however models struggled with this change.

\citet{musker2024semanticstructuremappingllmhuman} designed a symbolic analogy task that incorporated semantic information by creating analogies that are completed by taking sematic knowledge from words provided as \emph{a}, \emph{c}, \emph{e} terms etc., and converting them to non-semantic strings, which they refer to as a "semantic content" task. For example, provided pairs may include a list of animals mapped to symbols, such as \emph{horse}:*, \emph{cat}:*, \emph{spider}:! and so on where all mammals are mapped to an asterisk, *, and all non-mammals are mapped to an exclamation point, !. They found that smaller models performed much worse than humans, and therefore focused primarily on GPT-4 \cite{openai2024gpt4technicalreport}, Claude 3 \cite{claude}, and Llama-405B \cite{grattafiori2024llama3herdmodels}, as well as tested performance in humans. They found that Claude 3 achieved human performance on the semantic content task, while the other two models fell below human performance. However given that this is just one experimental setup, more investigation should be done in order to evaluate whether models can solve these sorts of analogies over a robust set of problems.

\subsection{Solving Entity-Level Analogies with Neural Networks}

\textbf{The Vector Offset Method.} Analogies are a popular benchmark to demonstrate that neural word embeddings encode knowledge beyond word similarity alone. The original method for solving analogies with Word2Vec- referred to as \emph{3CosAdd} by \citet{rogers-etal-2017-many}- involves finding the word, \textit{v}, in a vocabulary, \textit{V}, that is most similar to the vector resulting from the equation \textit{a+c-b=d'}, where \textit{a, b, c} and \textit{d}, are the vector representations of the terms in the analogy \textit{a:b::c:d} and \textit{d'} is the estimated vector representation for \textit{d} given the representations of \textit{a, b} and \textit{c}. This is to be compared to the actual, static vector representations $v \in V$ for \emph{d}estimated through training \cite{mikolov2013efficientestimationwordrepresentations, drozd-etal-2016-word, mikolov-etal-2013-linguistic}. Similarity in this case is defined as cosine similarity, \textit{cos(d, d')}. While the list of possible \emph{d} terms are finite in that they are limited to the model's predefined vocabulary size, this method still involves the {\fontsize{11pt}{11pt}\selectfont\textsc{retrieval}} process as all items (or most) in the vocabulary known to the model are potential candidates.

Researchers have identified several problems with this method of solving analogies. \citet{rogers-etal-2017-many} and \citet{linzen2016issuesevaluatingsemanticspaces} found that performance using 3CosAdd is heavily related to the proximity of \textit{a, b} and \textit{c} terms in semantic space, and that the ability for 3CosAdd to correctly identify analogies decreases as similarity between the entities in the relation decreases. This problem was found to hold with other vector offset methods presented in \citet{drozd-etal-2016-word} and \citet{levy-goldberg-2014-linguistic}. \citet{linzen2016issuesevaluatingsemanticspaces} also found that if you do not exclude \textit{a, b} and \textit{c} from \textit{V} as an estimate of \textit{d}, as was done in \citet{mikolov-etal-2013-linguistic} paper, that often the vector offset method will predict \emph{b} or \emph{c}.

Additionally, this method generally only allows options for \emph{d'} terms to be selected from a model's predefined vocabulary that was determined before training. \citet{chan:hal-03674913} address this problem by proposing a model that generates the \emph{d'} terms allowing for out-of-vocabulary generation, as opposed to retrieving them from a finite list. Their approach involved training a BiLSTM model to take as input each character of a word to encode a representation of the word. The 3CosAdd method is then applied to get a single representation of the analogy, and sent through a LSTM decoder to generate the \emph{d'} terms. They achieved almost perfect performance on morphological analogies from eight languages.

\textbf{Analogies for Model Probing.} Several papers have used semantic entity-level analogies as a probing tasks for LMs \cite{ushio-etal-2021-bert,czinczoll2022scientificcreativeanalogiespretrained, Chen_2022, yuan-etal-2023-beneath}, often while releasing an entity-level specific dataset. \citet{ushio-etal-2021-bert} and \citet{czinczoll2022scientificcreativeanalogiespretrained} both found that models generally perform worse on datasets comprised of only semantic-level analogies than on those that contain syntactic relations. When fine-tuning models with the BATS dataset both \citet{czinczoll2022scientificcreativeanalogiespretrained} and \citet{yuan-etal-2024-analogykb} found that performance is reduced on semantic analogy datasets, suggesting that these sorts of analogies are inherently different, and that semantic analogy datasets require the understanding of more abstract relations to solve.

\textbf{Algorithms for {\fontsize{11pt}{9pt}\selectfont\textsc{mapping}} Analogies.} Analogical {\fontsize{11pt}{11pt}\selectfont\textsc{mapping}} can additionally be performed between entities using algorithms that utilize information from LMs, without specifically performing {\fontsize{11pt}{11pt}\selectfont\textsc{mappings}} between the neural representations of entities themselves. \citet{jacob2023fameflexiblescalableanalogy} developed an algorithm to map between entities that utilizes LMs in several ways. First, they utilize GPT-3 as a knowledge base to extract relations between entities. They then use SBERT \cite{reimers2019sentencebertsentenceembeddingsusing} to calculate the similarity of relations between two systems, and to cluster relations within a system that are similar (the \emph{decomposition} method for {\fontsize{11pt}{11pt}\selectfont\textsc{rerepresentation}} presented in Section \ref{rerep}). Since the algorithm relies on similarities of relations between entities as opposed to similarity between entities specifically, they found that their algorithm may be more resilient to focusing on surface similarity between entities than focusing on similarities between the words themselves.

\textbf{Fine-tuning Models with Analogies.} \citet{yuan-etal-2024-analogykb}, \citet{Chen_2022}
 and \citet{petersen-van-der-plas-2023-language} tested whether analogy solving is something that can be learned with training on mapped analogies. \citet{yuan-etal-2024-analogykb} released a dataset of analogies created from knowledge graphs, (ConceptNet \cite{speer2018conceptnet55openmultilingual} and Wikidata \cite{vrandevcic2014wikidata}), and found that fine-tuning on their dataset improved analogy generation on out-of-domain semantic analogies, with T5-large \cite{10.5555/3455716.3455856} gaining 44-63 percentage points on test-sets after fine-tuning as compared to the vanilla model, reaching up to 80\% accuracy. On an analogy recognition task, they found that RoBERTa-large \cite{liu2019robertarobustlyoptimizedbert} and DeBERTa-v3 \cite{he2020deberta} trained on their data also often gained on performance.

\citet{petersen-van-der-plas-2023-language} attempted to address the geometry between word embeddings specifically by training models on analogies to maximize the cosine similarity between the differences of the entities, \emph{cos(a-b, c-d)}. Notably this is different from the vector offset method, which estimates a specific \emph{d} term using the equation \emph{cos(c-a+b, d')} and is generally a method for evaluation. They found that while training models to identify analogies when comparing differences between entities within the same domain (e.g. \emph{cos(a-b,c-d)}) improved performance on this task, detecting similarities between the longer distance connections (e.g. \emph{cos(a-c,b-d)}) did not improve with training.

\textbf{Model Prompting and Few Shot Learning.} Research in this area for entitly-level analogies can be split into two section- templates for non-causal LM's such as BERT to predict entities in a cloze-style test, and engineering prompts to elicit text generation.

\citet{ushio-etal-2021-distilling} introduce RelBERT, a RoBERTa model fine-tuned on triples formed from the SemEval-2012 task-2 \cite{jurgens-etal-2012-semeval} to generate relation embeddings between two entities. They build different RelBERT models with several prompting methods, testing both manual and learned prompts (AutoPrompt \cite{shin-etal-2020-autoprompt} and P-tuning \cite{liu2023gptunderstands}). While the model was trained to generate relation embeddings apart from the analogy setting, the authors test their model on its abilities to solve analogy datasets that were not seen during training. All analogies were multiple choice, therefore did not involve any {\fontsize{11pt}{11pt}\selectfont\textsc{retrieval}} process, only {\fontsize{11pt}{11pt}\selectfont\textsc{mapping}}, and did not require explicit verbalization of the relation embedding that was compared. Despite solving analogies in a zero-shot setting, RelBERT with manual prompting outperformed other baselines including few-shot GPT-3. All RelBERT models regardless of prompting choice outperformed few shot GPT-3.

When probing models with their novel analogy benchmark, ScAR, \citet{yuan-etal-2023-beneath} tested a variety of prompting methods to explore how background information and chain-of-thought (CoT) prompting \cite{10.5555/3600270.3602070} effect LM's reasoning abilities(in their case, large language models mostly with at least around 7B parameters such as InstructGPT). They found that including CoT prompting or providing background information improved robustness to prompt templates across 11 instruction designs. Additionally, they found that including background examples were particularly useful for the Chinese version of the ScAR across LM's, which they attribute to models' difficulties with Chinese domain-specific entities.

%\cite{brown2020language}, \cite{czinczoll2022scientificcreativeanalogiespretrained},\cite{rezaee-camacho-collados-2022-probing},\cite{yuan-etal-2024-analogykb}

\textbf{Model Scale.} When introducing \citet{brown2020language} GPT-3, the authors tested GPT-3 models of varying size, ranging from 125M to 175B parameters on the SAT dataset. They found that model size was correlated with performance, with the biggest gains with model size demonstrated in the few-shot prompting context, where the 125M parameter model achieving around 30\% accuracy and the 175B reaching 65\% accuracy. In the case of fine-tuning, when the dataset is small- as is often the case with analogy datasets- increased model size may not lead to increased performance, as demonstrated in \citet{petersen-van-der-plas-2023-language}, where BERT-base was able to improve accuracy on an analogy identification task using a relatively small training dataset, while BERT-large was unable to learn.

%\cite{yuan-etal-2023-beneath}

\textbf{Comparison to Human Performance.} Aside from the Google Analogy Testset and the BATS, there is no widely used benchmark to assess analogical reasoning for LM's. However, there is a decent-sized body of research on analogical reasoning performance in humans, often that provide the utilized datasets as well as estimates of human performance.

\citet{yuan-etal-2023-beneath} found, unsurprisingly, that larger models such as GPT-4 were able to match or exceed human performance on certain semantic and morphological benchmarks. However, when using the E-KAR dataset to test models and humans performance with regards to accuracy and a relational structure identification test- which tests the ability to correctly identify the relation involved in an analogy, they found that the overlap between being able to correctly solve an analogy and correctly identify the analogous relation was lower in models than in humans. The authors additionally tested domain transfer between analogies with a dataset they released (SCAR) and found that analogies between relatively similar yet still disparate domains saw higher accuracy in cross domain transfer than between more disparate domains. 
\citet{stevenson2023largelanguagemodelssolve} tested models' abilities to solve analogies and how they compared to the performance of children aged 7-12. They found that the LMs they tested outperformed 7 year-olds on an analogy task, and that several models such as RoBERTa and GPT-3 performed at the level of 11 year-olds. However after some investigation, they found that models may rely on associations between potential \emph{d} terms with the given \emph{c} to solve some analogies, as opposed to utilizing analogical reasoning by considering the \emph{a} and \emph{b} terms, suggesting similar issues as those presented with the vector offset method for Word2Vec despite these models being more sophisticated.

%\cite{petersen-van-der-plas-2023-language}

\subsection{Identifying Analogies in Context} \label{context}

\textbf{Mapping at the Entity Level within Context.} \citet{sultan2023lifecircusclownsautomatically} took an algorithmic approach to {\fontsize{11pt}{11pt}\selectfont\textsc{mapping}} entities that are analogous between procedural texts, which they call Finding Mapping by Question. They use QA-SRL \cite{fitzgerald2018largescaleqasrlparsing} to generate questions and answers regarding the sentences present in the procedural texts. The questions function as a way to identify similar entities between texts, with the assumption that similar entities would have similar questions. The authors also used a clustering algorithm to address coreference issues in the texts, which, much like the \citet{jacob2023fameflexiblescalableanalogy} paper, could be considered \emph{decomposition}. Beam search was used to finalize {\fontsize{11pt}{11pt}\selectfont\textsc{mapping}} between systems.

\textbf{Higher Level Analogical Abstraction.}
Some research has attempted to identify analogies between two texts "in general", i.e. no specific entities or relations contained in the text are mapped, but instead multiple texts are deemed analogous at a higher level. For example, \citet{ghosh2022epicemployingproverbscontext} released ePiC,
a dataset of narratives associated with various
proverbs, and tested whether models could predict the corresponding proverb, with the analogy
being between the moral of the narrative and the
moral of the proverb. Additionally, \citet{nagarajah2022understandingnarrativesdimensionsanalogy, sourati2024arnanalogicalreasoningnarratives} and \citet{jiayang2023storyanalogyderivingstorylevelanalogies} try to identify the presence of analogies between entities or relations in texts, but do not specifically identify and map them. This is perhaps more relevant to the {\fontsize{11pt}{11pt}\selectfont\textsc{abstraction}} and {\fontsize{11pt}{11pt}\selectfont\textsc{encoding}} processes than performing {\fontsize{11pt}{11pt}\selectfont\textsc{mapping}}, since they identify whether the overall structure has similarities.

\textbf{Contextual Analogy Generation.} Another task that has been attempted is that of analogy generation- formulated as providing a source concept for which the generated text should be analogous \cite{bhavya2022analogygenerationpromptinglarge,ding2023fluidtransformerscreativeanalogies,sultan2024parallelparcscalablepipelinegenerating}. Like entity-level generation, this task is related to the {\fontsize{11pt}{11pt}\selectfont\textsc{retrieval}} process, as any generated analogy would be based on knowledge the model contains. 

\textbf{Fine-Tuning.}
Unfortunately, given the relatively small size of contextual-analogy datasets (until recently), experiments exploring fine-tuning on these sorts of analogies has been minimal and not particularly informative. \citet{jiayang2023storyanalogyderivingstorylevelanalogies} addressed the lack of data issue, and introduced the StoryAnalogy dataset, which includes 24K contextual analogy pairs. On an  Semantic Textual Similarity (STS) style analogy identification task, they found that fine-tuning RoBERTa models improves classification ability on a style task, and that fine-tuned models performed better on their novel analogy evaluation metric than larger LLMs such as LLaMa-65B, but overall correlation with human scores had room for improvement. They also found that a fine-tuning FlanT5 models \cite{chung2024scaling} improved analogy generation and the novelty of the generations over the model in the few-shot setting, but that plausibility decreased with tuning.%\citet{nagarajah2022understandingnarrativesdimensionsanalogy} tuned DistilRoBERTa-base \cite{sanh2020distilbertdistilledversionbert} 
 %on two tasks- an analogy type prediction task focused on identifying the presence of type of analogy between two story pairs (please see original paper for the categories), as well as an analogical transfer learning task. They found that the F1 score was high for types that had an equal distribution of positive and negative instances, but poorly for those with skewed distribution. The authors state that they believe the model would perform better in the case of more data.

\textbf{Model Prompting.}
\citet{sultan2024parallelparcscalablepipelinegenerating} introduced a data generation pipeline using GPT-3.5, ParallelParc, to generate contextual analogies. They found that prompting GPT-3.5 to generate analogies with no guidance resulted in analogies that were repetitive or between similar topics. After including the  the base system for which GPT-3.5 needed to create an analogical paragraph in a single prompt, the model tended to generate paragraphs for target systems that differed mostly through changing nouns of the base system. They ultimately found that a two-step prompt, one that first identified an appropriate target concept and the analogous relations, and a second that generated the text, achieved the desired results.

Among humans, it is common for teachers to use analogies to explain a newly introduced or more unfamiliar concept. \citet{yuan-etal-2024-boosting} tested this in the LM setting, where analogies, including long-form analogies, generated by LM's where used as prompts that included background information on two scientific question answering datasets for student models. They found that including long-form analogies in the prompt outperformed zero-shot and CoT prompting.

%\cite{sourati2024arnanalogicalreasoningnarratives}
%\cite{jiayang2023storyanalogyderivingstorylevelanalogies}

\textbf{Model Scale.}
On their StoryAnalogy dataset, \citet{jiayang2023storyanalogyderivingstorylevelanalogies} found that larger models like GPT-3.5 and ChatGPT did not outperform smaller encoder models on a the STS style tasks. \citet{combs2025zero} tested 13 large models -the smallest being StableLM \cite{bellagente2024stablelm216b} with 1.6B parameters and the largest with known number of parameters being GPT-4 with 1.8T- on datasets taken from \citet{gentner1993roles} and \citet{wharton1994below}. They found that while GPT-4, GPT-4o, and Claude were consistently among the top performers in the experiments, model performance was not completely correlated with size.

\textbf{Comparisons to Human Performance.} \citet{sourati2024arnanalogicalreasoningnarratives} tested the ability of six models, which were instructed to choose the narrative that was analogous to a given source narrative, and were presented with different combinations of near/far analogies and near/far disanalogies. They also tested SBERT, using cosine similarity between the two narratives to determine whether the pairs were analogous. This task was formulated as a binary analogy selection task, and as mentioned before, arguably addresses {\fontsize{11pt}{11pt}\selectfont\textsc{abstraction}} and {\fontsize{11pt}{11pt}\selectfont\textsc{encoding}} instead of {\fontsize{11pt}{11pt}\selectfont\textsc{mapping}}.

While none of the models were able to match human performance, GPT-4 was able to approach human performance when the true analogy option was a near analogy. On average, models performed better when the true analogy was near and the disanalogy was far, and the worst when the true analogy was far, and the distractor was near.

%On the Semantic Textual Similarity (STS) style analogy identification task %\citet{jiayang2023storyanalogyderivingstorylevelanalogies}
%\cite{sultan2024parallelparcscalablepipelinegenerating}

\section{Applications of Analogies}\label{uses} % Going beyond solving analogies % Going beyond analogy solving

Analogical reasoning is not just an exercise in relational thinking performed for fun or used as a broad test of intelligence, it is a tangible way to formulate many domain-specific tasks. A perhaps obvious one would be for creative processes, such as metaphor generation in creative writing \cite{gentner2001metaphor}.

Additionally analogy is often used in education, from early education to medical school \cite{heywood2002place, guerra2011analogies, pena2010analogies,gray2021teaching}. It also has application in health and medicine  \cite{alsaidi2022analogy,guallart2014analogical}, science communication \cite{schwarz2018nanotechnology,corner2015like,elliott2016communicating}, innovation and creative problem solving \cite{gick1980analogical,innovation, markman2009supporting}, and law \cite{sep-legal-reas-prec,condello2016metaphor}.

\subsection{Broader Incorporation of Analogy and Analogical Thinking in NLP}  % maybe make shorter

We would like to discuss potential avenues for incorporating analogical reasoning in NLP at large. Analogical reasoning can potentially address certain known issues in NLP, such as spurious correlations, bias, out of domain generalization, and explainability.

When discussing different categories of explainability methods, \citet{lyu2024faithfulmodelexplanationnlp} cover similarity-based methods, where similar previous examples are used to justify why a model made a decision given the current input. They describe these methods as being similar to how humans use analogy. Identifying relevant previous examples is related to the {\fontsize{11pt}{11pt}\selectfont\textsc{retrieval}} process of analogical reasoning, however, these methods can be subject to spurious correlation, perhaps much like the {\fontsize{11pt}{11pt}\selectfont\textsc{retrieval}} processes in humans can be effected by surface similarity. Additionally, much like the research with contextual analogies detailed in section \ref{context}, these methods do not always map between specific spans of text. One way to improve the reliability and transparency of similarity-based methods would be to incorporate a {\fontsize{11pt}{11pt}\selectfont\textsc{mapping}} process as opposed to just a {\fontsize{11pt}{11pt}\selectfont\textsc{retrieval}} process, where specific entities, concepts, or themes in the retrieved text can be mapped to the target text. \citet{Ming_2019} introduced an explainability method, ProSeNet, which learns prototypical examples that are then used to explain predictions on subsequent data, highlighting relevant text. However, the prototypical examples learned are not necessarily actual examples present in the dataset, and are not retrieved on a case by case basis. This might be relevant for some applications that require more specificity, such as identifying similar ruling for individual court rulings in the law domain.

Analogical reasoning could also be used to address out-of-domain generalization and transfer learning. Transfer of knowledge or expertise in humans is an area of cognitive science that is heavily investigated, often in the context of analogical reasoning, and generally involves the {\fontsize{11pt}{11pt}\selectfont\textsc{encoding}} and {\fontsize{11pt}{11pt}\selectfont\textsc{retrieval}} processes \cite{kimball2000transfer}. In order to transfer knowledge, one must be able to retrieve the relevant instance in memory. The ability to both retrieve and then transport the solution to a new situation depends on how both the original and target instances are encoded  \cite{gick1983schema,loewenstein2010one}. Understanding text not merely as a sum of its parts, but as representing ideas and solutions that can be generalized to alternate domains or situations, and allowing these representations to be updated or adapted when presented with new information, is arguably vital to cross-domain generalization \cite{Doumas_2022}. Furthermore, there may be domains that have a scarce availability of training data, where being able to identify generalizable, domain agnostic abstractions could be beneficial. This does not have to be limited to the training or evaluation stage, for example, \citet{huang-etal-2024-araida} presented Analogical Reasoning-Augmented Interactive Data Annotation (ARAIDA), to address the annotation stage.

A known problem in NLP is the tendency of models to rely on spurious correlations, where certain words or terms are heavily associated with a label despite being irrelevant to the task at hand \cite{wang-culotta-2020-identifying}. This can be thought of as the model relying on surface similarity to perform tasks. In order to overcome this, models need to learn to identify and attend to more abstract and relational information and higher level perceptual information contained in text \cite{chalmers1992high}.

Somewhat related is the relevance of multi-model input, situational learning, and incorporating our various human senses and perceptual capabilities into processing and understanding language data, which are ways in which human learning and machine learning fundamentally differ \cite{frank2023bridging, beuls2024humans}. Arguably all the processes involved in analogical reasoning (and cognitive processes in general) in humans involve incorporating perceptual data \cite{Mitchell_2021}. \citet{kotovsky2013representation} and \citet{zamani2000object} suggested that how a concept is encoded is important to a cognitive system's ability to recognize analogy with visual stimuli. For humans, context in language is not limited purely to distributional semantics. Language takes place in the context of all our senses and interactions. With that said, this can potentially be a double edged sword. While additional, non-textual information is influential in helping us understand language, there is also the issue mentioned in Section \ref{abtract}, where external stimuli can also be distracting for quality {\fontsize{11pt}{11pt}\selectfont\textsc{abstraction}} of ideas.

Lastly, current NLP tasks could be reframed in analogical way, which has already been done in the literature to some extent. For example, \citet{wang2020vector} formulated sentence generation as a sentence analogy task, where the desired sentences to be generated \emph{d} were edited versions of sentences \emph{c}, where the necessary edits were the differences between two other given sentences \emph{a} and \emph{b}. 
\citet{wijesiriwardene2024relationshipsentenceanalogyidentification, thilinilevels} reformulated natural language inference, specifically the entailment and negation labels, as solving analogy problems. One could also argue that coreference resolution is related to the {\fontsize{11pt}{11pt}\selectfont\textsc{rerepresentation}} process, and that the methods outlined in \citet{yan2013theory} could be applied to approaches addressing coreference resolution to tackle other tasks such as natural language inference.

\section{Conclusion and Limitations}

In this paper, we review the processes of analogical reasoning in humans and connect them with current research and methods in NLP. We found that experiments with certain types of analogies typically focus on specific processes (ex: entity level and mapping). Additionally, we also suggest that analogical reasoning could potentially be an approach to address current limitations of NLP models, and be a way for researchers to focus more on optimizing relational understanding and similarity rather than relying on entity similarity.

One major limitation of this work is we just brushed the surface of research in cognitive science regarding analogical reasoning. Other areas of analogical reasoning that could also be of interest to the NLP community which were not touched upon would be the development of analogical and relational reasoning over the humans lifespan and how this could help improve analogical transfer.

Additionally, we did not address research with knowledge graphs, relation embeddings, and other similar areas of research which are often incorporated with NLP to create more knowledge-rich representation. However, given the limitations of manually crafted knowledge bases, such as being ultimately limited in scope and time-intensive to build \cite{schwartz2009acquiring,yuan-etal-2024-analogykb}, being able to understand and extract relations in text can ultimately not depend on them.

\section{Acknowledgments}
We are grateful to the Swiss National Science Foundation (grant 205121\_207437 : C-LING) and the Fondazione Aldo e Cele Daccò for funding this work. We also thank members of the Idiap NLU-CCL group for helpful discussions, and the anonymous reviewers for their fruitful comments and suggestions.

% Submission-specific rules

\bibliography{tacl2021}
\bibliographystyle{acl_natbib}

\iftaclpubformat

\onecolumn

\fi

\end{document}